\documentclass{llncs}
\usepackage{algorithmic}
\usepackage[boxed,ruled,lined,linesnumbered,longend]{algorithm2e}
\usepackage{amssymb}
\usepackage{amsmath}
\usepackage{epsfig}
\usepackage{graphicx}
\usepackage[ansinew]{inputenc} 
\usepackage{subfigure}

\begin{document}

\title{Design of Emergent and Adaptive Virtual Players in a War RTS Game}

\titlerunning{}  
%
\author{José A. Garc\'{\i}a Guti\'{e}rrez \and Carlos Cotta \and Antonio~J.~Fern{\'a}ndez Leiva}
\authorrunning{J.A. ~Garc\'{\i}a \and C.~Cotta \and A.J.~Fern{\'a}ndez}   
\institute{Dept.~Lenguajes y Ciencias de la Computaci{\'o}n, ETSI Inform{\'a}tica, \\
Campus de Teatinos, Universidad de M{\'a}laga, \\
29071 M{\'a}laga -- Spain \\
\email{\{ccottap,afdez\}@lcc.uma.es}}

\maketitle

\begin{abstract}
Basically, in (one-player) war Real Time Strategy (wRTS) games
a human player controls, in real time, an army consisting of a
number of soldiers and her aim is to destroy the opponent's assets
where the opponent is a virtual (i.e., non-human player controlled)
player that usually consists of a pre-programmed decision-making
script. These scripts have usually associated some well-known
problems (e.g., predictability, non-rationality, repetitive
behaviors, and sensation of artificial stupidity among others). This
paper describes a method for the automatic generation of virtual
players that adapt to the player skills; this is done by building
initially a model of the player behavior in real time during the
game, and further evolving the virtual player via this model
in-between two games. The paper also shows preliminary results
obtained on a one-player wRTS game constructed specifically for
experimentation.
\end{abstract}

%

\vspace{-2mm}
\section{Introduction and related work}


\begin{sloppypar}
In an era in which computing power has boosted the graphical quality
of videogames, players have turned their attention to other aspects
of the game. In particular, they mostly request opponents exhibiting
intelligent behavior. However, intelligence does not generally
equate to playing proficiency but rather to interesting behaviors
\cite{liden:artifical-stupidity-game-wisdom2-2004}. This issue is
specifically relevant in real-time strategy (RTS) games that often
employ two kinds of artificial intelligence (AI)
\cite{ahlquist+:game_ai08}: one represented by a {\em virtual
player} (VP, or non-player character -- NPC) making decisions on a
set of
units (i.e., warriors, 
machines, etc.), and another one corresponding to the small units
(usually with little or no intelligence).
The design of these AIs is a complex task, and the reality in the
industry is that in most of the RTS games, the NPC is basically
controlled by a fixed script that has been previously programmed
based on the experience of the designer/programmer. This script
often comprises hundreds of rules, in the form {\sc if the game is
in state S then Unit U should execute action A}, to control the
behavior of the components (i.e., units) under its control. This
arises known problems: for instance, for players, the opponent
behavior guided by fixed scripts can become predictable for the
experienced player.
Also, for AI programmers, the design of virtual players (VPs) can be
frustrating because the games are becoming more and more complex
with hundreds of different situations/states and therefore it is not
easy to predict all the possible situations that could potentially
happen and even more difficult to decide which is the most
appropriate  actions to take in such situations. As consequence,
many RTS games contain 'holes' in the sense that the game stagnates
or behaves incorrectly under very specific conditions (these
problems rely on the category of 'artificial stupidity'
\cite{liden:artifical-stupidity-game-wisdom2-2004}). Thus the
reality of the simulation is drastically reduced and so too the
interest of the player;
\end{sloppypar}

In addition, there are very interesting research problems in
developing AI for Real-Time Strategy (RTS) games including planning
in an uncertain world with incomplete information, learning,
opponent modeling, and spatial and temporal reasoning
\cite{buro:AAAI2004}. This AI design is usually very hard due to the
complexity of the search space (states describe large playing
scenarios with hundreds of units simultaneously acting). Particular
problems are caused by the large search spaces (environments
consisting of many thousands of possible positions for each of
hundreds, possibly thousands, of units) and the parallel nature of
the problem - unlike traditional games, any number of moves may be
made simultaneously \cite{DBLP:conf/gameon/CorrubleMR02}.
Qualitative spatial reasoning (QSR) techniques can be used to reduce
complex spatial states (e.g., using abstract representations of the
space \cite{DBLP:journals/expert/ForbusMD02}). Regarding
evolutionary techniques,  a number of biologically-inspired
algorithms and multi-agent based methods have already been applied
to handle many of the mentioned problems in the implementation of
RTS games
\cite{DBLP:journals/tec/LouisM05,DBLP:journals/tec/StanleyBM05,DBLP:conf/cig/Livingstone05,miles+:co-evol-rts2006,DBLP:conf/evoW/LichockiKJ09,beume+:cig2008,keaveney+:cig2009,DBLP:conf/aiide/HagelbackJ09}.

Even in the case of designing a very good script for VPs, the
designer has to confront another well known problem: the VP behavior
is usually fixed and rarely adapts to the level (i.e., skills) of
the player. In fact the player can lose interest in the game because
either she is able to beat all the opponents in each level, or the
virtual player always beat her.
In this sense, the design of interesting (e.g., non-predictable)
NPCs is not the only challenge though. It also has to adapt to the
human player, since she might lose interest in the game otherwise
(i.e., if the NPCs are too easy or too hard to beat).

There are many benefits attempting to build adaptive learning AI
systems which may exist at multiple levels of the game hierarchy,
and which evolve over time. This paper precisely deals with the
issue of generating behaviors for the virtual player that evolve in
accordance with the player's increasing abilities. This behavior
emerges according to the player skill, and this emergent feature can
make a RTS game more entertaining and less predictable in the sense
that emergent behavior is not explicitly programmed but simply
happens \cite{sweetser:emergence_in_games2008}. The attainment of
adjustable and emergent virtual players consists here of a process
of two stages that are iteratively executed in sequence: (1) a
behavior model of the human player is created in real time during
the execution of a game, and further the virtual player is evolved
off-line (i.e., in between two games) via evolutionary algorithms
till a state that it can compete with the player (not necessarily to
beat her but to keep her interest in the game). This approach has
been applied on a RTS game constructed specifically for
experimentation, and we report here our experience.

\vspace{-2mm}
\section{The Game}
\label{sect:the game}

Here we describe the basic elements that compose our wRTS game.

{\em Scenario}. It will also be called indistinctly {\em region} or
{\em world} and consists of a two-dimensional non-toroidal
heterogeneous hostile and dynamic grid-world. The world is {\em
heterogeneous} because the terrain is not uniform, {\em hostile}
because there exist a virtual army whose mission is to destroy the
human player-controlled army, and {\em dynamic} because the game
conditions change depending on the actions taken by units of both
armies. Regarding the heterogeneity of the terrain, each grid in the
region can have one of the three following values: passable (each
uni can traverse this grid), impassable (no unit can traverse it),
and semi-impassable (there is a penalization of 1 point of energy -
see below -to traverse it).

{\em Army}. There are two armies (also called indistinctly teams,
represented by spiders and ladybirds) with a number of units (also
called indistinctly soldiers or agents) and a flag to be defended.
One army is controlled by the human player whereas the other one is
guided by some kind of artificial intelligence (AI).

Given a unit $u$ placed in a position $(x_u,y_u)$ in the grid, its
{\em visual range} (VR$_u$) embraces any position $(x,y)$ that is
placed to a maximum distance $\phi$, that is to say, $\text{VR}_u =
\{(x,y) \mid \sqrt{(x_u-x)^2+(y_u-y)^2} \leq \phi\}$. Initially,
each agent only knows the grids that belongs to its visual range. In
fact, during the game the scenario is not completely known for an
army and only those regions that were already visualized by its
agents are known. The global information that each army
(independently if this is controlled by the AI or the player) has is
the sum of all the information of their constituent soldiers. Note
that (human or virtual) players only know the rival flag position if
this has been detected in a grid by some of its agents previously.
Also, each unit interacts with the environment by executing a number
of actions (see below) and has certain level of health and energy
that decreases with these interactions. The initial values for the
health and energy of each soldier are 100 and 1000 respectively.

{\em Fights.} Full body combat can be executed at the soldier level
by assigning a random value (in the interval [0,1]) to each unit
involved in a fight with a rival soldier; the energy level of the
unit with lowest value decreases 1 unit. A unit dies when its energy
is zero. All the combats between soldiers are executed in sequence
in one turn of the game.

{\em Decision-making.} Along the game, the role of players is to
make decisions with respect to the actions that the units of their
respective associated armies should take. More specifically, both
the virtual and human player can select a set of agents and further
assign it a specific order that basically consists of a common
action that each unit in this set has to execute.

{\em Actions.} The game is executed in turns; every turn, each
soldier executes the last order received by its player. Six actions
are  possible in this game: \vspace{-3mm}
\begin{itemize}
\item Move forward enemy: if applied to unit $u$, then this has to move towards the closer rival soldier according
to $\text{VR}_u$, otherwise (i.e., if no rival soldier exists in
$\text{VR}_{u}$) agent $u$ has to move in direction to the region
where there are more enemies according to the global information of
the army.

\item Group/Run away: the soldier has to group with other team mates placed
inside its visual range; if there are not mates in this range, then
the agent has to move towards the position where its flag is placed.

\item Move forward objective: the agent advances towards the opponent's flag position if this is
known, otherwise it moves randomly.

\item No operation: execute no operation.

\item Explore: the soldier moves towards a non-explored-yet region of the
map.

\item Protect flag: The soldier has to move towards the position of its own flag; if it is already near to it, unit
just should scout the zone.
\end{itemize}

{\em Perceptions}. Each unit knows its specific situation in the
game from the perceptions that receives in its visual range.
In our game, the {\em agent state} is determined by its health
(measured in three range values: low, medium and high), and the
following boolean perceptions:
\begin{enumerate}
\item Advantage state? ($S_a$): ``the unit is into a handicap situation". This  perception is true if, according to the information in its visual range,
the number of mate soldiers is higher than the number of rival
units, and false otherwise.

\item Under attack? ($U_a$): this perception is true of the soldier is
involved in a full body combat with another opponent unit, and false
otherwise.

\item Objective visible? ($O_v$): the opponent's flag is visible in its visual range.
\end{enumerate}

{\em Game objective}. The army that first captures the rival flag is
the winner; if this situation is never reached after a number of
turns, then the winner is the player whose army inflicts a higher
damage to its rival army and where the damage is measured as number
of dead units.

\subsection{Notes on specific issues}

Observe that `fighting' is not a specific action to be executed by
the army units. This should not be surprising because `fighting' is
not usually controlled by players in standard RTS games as this
action should often is executed at the unit level and not at the
army level; in our game this action is automatically executed when a
unit and a rival soldier meet in any grid of the scenario.

Also, note that five of the six possible actions that units can
execute require to make movements in the game world. This basically
means that some kind of pathfinding should be processed to do the
movement realistic. Classical algorithms such as A$^{*}$ were not
good candidates for this task since, as already indicated, most of
the scenario is not known for the army units. A practical solution
was to let units make a good-enough movement depending on its
scenario knowledge; several cases were covered: {\em Concave
obstacles}: the unit moves by considering different angles with
respect to its position in the scenario; {\em Convex obstacles}: the
unit moves by following the obstacle contour in a specific
direction; and {\em Bridges and narrow passages}: a simple
ant-colony algorithm was executed.

\begin{figure}[htb]
\vspace{-8mm} \centering \subfigure[Concave obstacles]{
    \label{fig:concave obstacle movement}
    \includegraphics[width=3.75cm,bb=0 0 506 378]{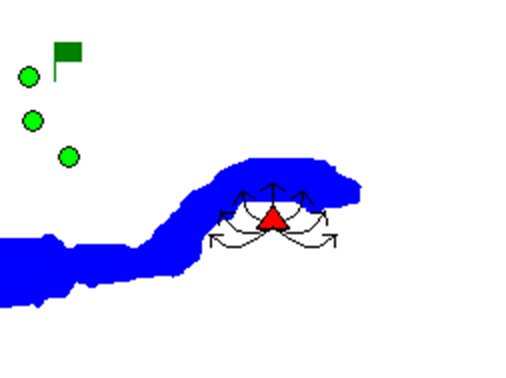}
} \subfigure[Convex obstacles]{
    \label{fig:convexe obstacle movement}
    \includegraphics[width=3.75cm, bb =0 0 312 378]{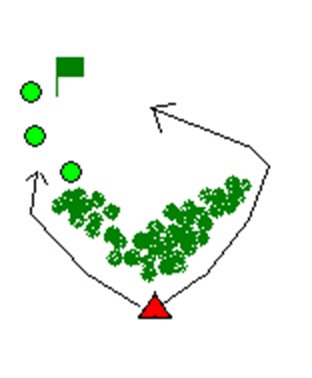}
} \subfigure[bridges and passages]{
    \label{fig:bridge obstacle movement}
    \includegraphics[width=3.75cm, bb = 0 0 354 380]{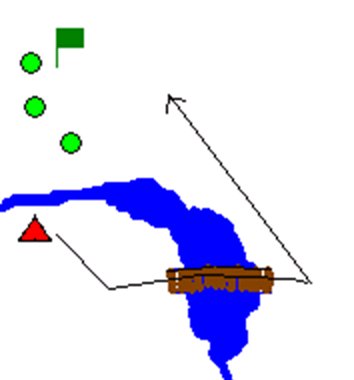}
} \subfigure[Ant-colony]{
    \label{fig:ant colony}
    \includegraphics[width=4.00cm, bb = 0 0 516 382]{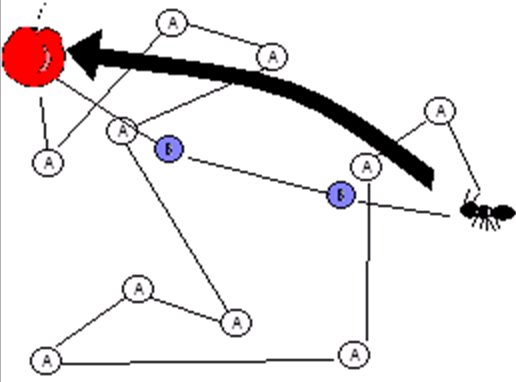}
} \subfigure[Game]{
    \label{fig:tuning1:e}
    \includegraphics[width=0.80cm,bb=0 0 167 156]{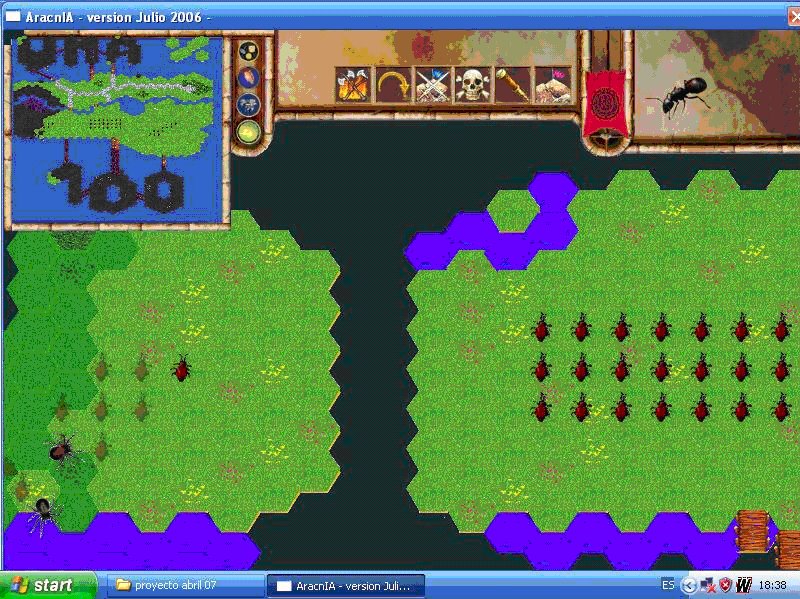}
} \caption{(a)-(c) Movement examples; (d) illustration of classical
execution of ant-colony algorithm used for pathfinding; (e) game
screenshot (black cells represent non-explored regions yet, blue
cells impassable grids, and green cells passable regions in the
scenario.} \label{Fig:movement}
\end{figure}

Due to space limitations we will not explain them in details as this
is beyond the objective of this paper. In any case, Figures
\ref{Fig:movement}(a),(b) and (c) show illustrative examples of the
cases, Figure \ref{Fig:movement}(d) illustrates the classical
behavior of our ant colony algorithms, and Figure
\ref{Fig:movement}(e) displays a screenshot of our wRTS
game\footnote{A prototype version of this game can be unloaded from
{\tt http://www.lcc.uma.es/$\sim$afdez/aracnia}.} titled {\em
aracnia}. Note also that a description of the technical issues of
the game (e.g., implementation of basic actions, graphical
considerations, input/output interface, scene rendering and memory
management, among others) is beyond the scope of this paper. In the
following we will focus  on the process of designing virtual
players.

\vspace{-10mm}
\section{Virtual player design}

Here we describe our proposal to design adaptive virtual players in
our wRTS.

\vspace{-3mm}
\subsection{General issues}

Our aim is to generate controllers governing automatically (i.e.,
without human intervention) the behavior of an entire team (i.e., a
set of autonomous agents/soldiers). An army controller can be viewed
as a set of rules that will control the reactions of its agents
according to the current state of play.
Depending on its situation, the
agent executes a particular action that may modify the state of the
agent itself. The global team strategy emerges as the result of the
sum of the specific behavior of all the agents.

However, the definition of specific strategies for each agent is
complex, requires  a profound knowledge of the required behavior of
each agent, and results in predictable behavior. Moreover, devising
independent controllers for each unit in a RTS game is costly; in
fact this is not realistic as requires a high computation effort
that surely decreases the quality of the real-time rendering of the
game. One solution consists of designing global team strategies;
however, again this is very complex as the programmer has to cope
with too many possibilities that arise from the different
interactions among all
 the agents of both armies. We have
opted by a more realistic process:  all the soldiers will be managed
by the same controller, and in any instant of the game, the team
behavior will be the result of summing all the action taken by each
of its constituent units. This means that we have to produce just
one controller to devise a global team strategy. Note however that
this does not mean  all the agents execute the same action because
the action to be executed by a soldier will depend on its particular
situation in the match.

\vspace{-5mm}
\subsection{A two phases process}

The procedure described in the following has the aim of creating a
virtual player whose behavior auto-evolves according to the skills
that human player exhibits along the game.
Algorithm~\ref{code:improvement} displays the general schema of the
process: initially (in the first game), the virtual player (VP) is
constructed as an expert system (i.e., a rule-based prototype (RBP))
that comprises tens of rules trying to cover all the possible
situations in which an agent might be involved. This system was the
result of a number of trials designed from the experiences of the
authors playing wRTS games. Then, let $\mathbb{N}_{h}$ be
$\{1,\ldots,h\}$ henceforth, and assume the human player will play a
number of $\wp$ games, the procedure consists of two phases that are
sequentially executed. Firstly, a {\em player modeling phase} is
conducted; this step is described in Section \ref{sect:user
modelling} and basically consists of building a model of the
behavior that human player exhibits during the game (i.e., on-line).
The second phase, described in Section \ref{sect:GA evolution}, is
devoted to construct the virtual player by evolving a strategy
controlling the functioning of a virtual army unit via an
evolutionary algorithm (EA). In the following we describe in details
both phases. Firstly, we discuss how the virtual player is
internally encoded  as this is important to understand the
functioning of the whole schema.

\begin{algorithm}[!t]
\scriptsize{ \caption{PMEA (VP)}\label{code:improvement} VP
$\leftarrow$ \textsc{RBP};\ \quad{\tcp{Rule-based expert system}}
\For{$i\in\mathbb{N}_{\text{$\wp$}}$}{  
   PlayerModel $\leftarrow$ \textsc{PlayGameOn}(VP);\ \quad{\tcp{Player modeling (PM)}}
   VP $\leftarrow$ \textsc{EA}(PlayerModel,VP);\ \quad{\tcp{Evolutionary optimization (EA)}}
} \Return{{\rm VP}}\; }
\end{algorithm}

\begin{figure}[t!]
\centerline{\includegraphics[width=5cm]{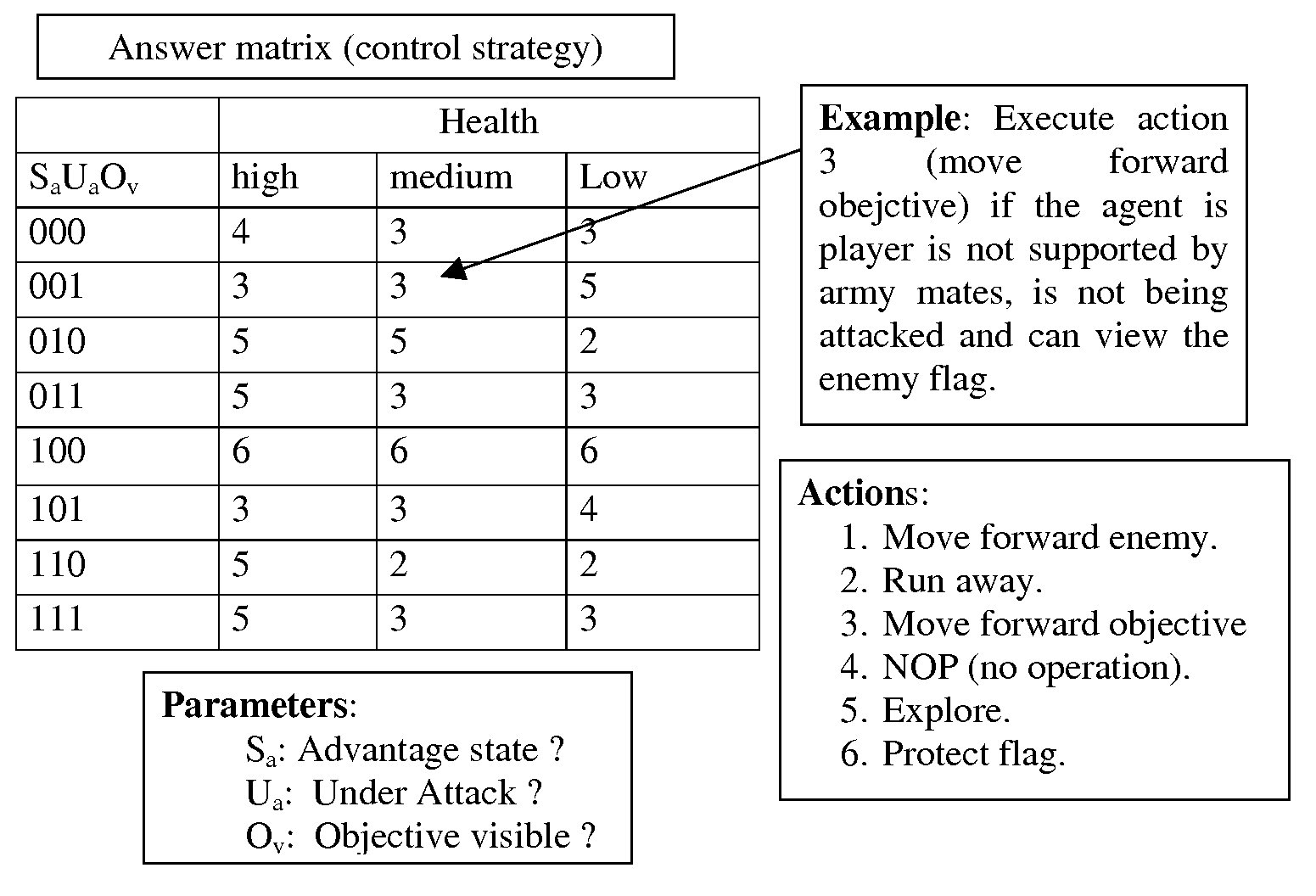}\includegraphics[width=5cm]{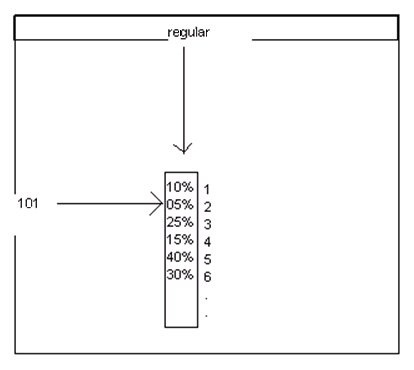}}
\caption{Example of an arbitrary encoding (left), and extended
answer matrix (right).} \label{fig:representation-data modelling}
\end{figure}

\begin{sloppypar}
{\bf Representation:} the virtual player contains a NPC army
strategy, and its internal encoding corresponds with both the
rule-based expert system and any individual in the EA population. In
the following we will refer to `individual' in a general sense.
Every individual represents a team strategy, that is to say, a
common set of actions that each army unit has to execute under its
specific situation in the environment. An individual is represented
as a vector $v$ of $k$ cells where $k$ is the number of different
situations in which the agent can be in the game, and $v[i]$ (for $0
\leq i < k$) contains the action to be taken in situation $i$. In
other words, assuming $m$ perceptions, where the perception $p_j$
(for $0 \leq j \leq m-1$) can have $k_j$ possible values, then $k =
k_0 * k_1 * \ldots * k_{m-1}$ and the cell:
$$
v[e_{m-1} + e_{m-2}*k_{m-1} + e_{m-3}
* (k_{m-1} * k_{m-2}) + e_{m-4} * (k_{m-1} * k_{m-2} * k_{m-3})+\ldots…]
$$
contains the action to be executed by a specific unit when its
perceptions $p_0, p_1,\ldots, p_{m-1}$ have the values $e_0,
e_1,\ldots, e_{m-1}$ respectively. As indicated in Section
\ref{sect:the game}, the state of a unit in our wRTS game is
determined by three boolean perceptions and its energy level (with 3
different values) so that the encoding of an individual consists of
a vector with 24 genes; this vector will be called {\em answer
matrix}, and each cell in the vector will contain an action.
Remember that 6 possible actions can be executed in our wRTS and
thus the search space (i.e., the number of different strategies that
can be generated from this representation) is $6^{24}$. Figure
\ref{fig:representation-data modelling}(left) displays an example of
a possible encoding. The optimal solution (if it exists) would be
that strategy which always select the best action to be executed for
the agents under all possible environmental conditions. In fact,
this vast search space makes this problem impracticable for exact
methods and justifies the use of evolutionary algorithms.
\end{sloppypar}

\subsection{On-line phase: User modeling}
\label{sect:user modelling}


The player behavior will be modeled as a ruled-based strategy
encoded as explained above. This process requires collecting, during
the execution of the game (i.e., in real time) the actions that
human player executes, recording additionally the specific
conditions under which these are performed. At the end of this
process we generate an {\em extended answer matrix ($v^+$)} that is
an answer matrix (i.e., an individual encoding $v$) where each cell
$v^+[i]$ (for $0 \leq i < k$) represents now a vector of 6 positions
(one per action) and $v^+[i][a]$ (for some action $a \in [1,6]$)
contains the probability that human player executes action $a$ under
the environment conditions (i.e., the state) associated to cell
$v[i]$. Figure~\ref{fig:representation-data modelling}(right)
displays an example that shows the probability of executing each of
the 6 possible actions in a specific situation of the unit (i.e.,
the soldier has medium energy, is in an advantage state, is not
suffering an attack, and knows where the opponent flag is placed).
This extended answer matrix is finally used  to design the virtual
player as follows: $\text{VP}[i] = argmax_{a \in [1,6]} \{v^+[i][a]
\}$, for all possible situations $i$.

\vspace{-3mm}
\subsection{Off-line phase: evolutionary optimization}
\label{sect:GA evolution}

Algorithm~\ref{code:ga} shows the basic schema of our EA. The
initial population is randomly generated except one individual that
is charged with the virtual player (lines 1-4). Then a classical
process of evolution (lines 6-21) is performed and the best solution
in the population is finally returned (draws are broken randomly).

\begin{algorithm}[!t]
\scriptsize
 \caption{EA(PModel, VP))\quad //\ PModel = Player Model, VP = Virtual Player}\label{code:ga}
\For{$i\in\mathbb{N}_{\text{popsize}-1}$} {
   $pop_i\leftarrow$\textsc{Random-Solution()}\;
} $pop_\text{popsize}\leftarrow$\textsc{VP}\;  $i\leftarrow0$\;
\While{$i < MaxGenerations$} {
    \textsc{Rank-Population} ($pop$); \tcp{sort population according to fitness}
    $parent_1\leftarrow$\textsc{Select} ($pop$); \tcp{roulette wheel}
    $parent_2\leftarrow$ \textsc{Select} ($pop$)\;
    \uIf(\texttt{// recombination is done}){${\rm Rand}[0,1]<p_X$}
    {
       $(child_1,child_2)\leftarrow$ \textsc{Recombine} ($parent_1, parent_2$)\;
    }
    \Else
    {
       $(child_1,child_2)\leftarrow (parent_1,parent_2)$\;
    }
    $child_1\leftarrow$ \textsc{Mutate} ($child_1, p_M$); \tcp{$p_M$ = mutation probability}
    $child_2\leftarrow$ \textsc{Mutate} ($child_2, p_M$)\;
    $fitness_1 \leftarrow \textsc{PlayGameOff}(PModel,child_1)$\;
    $fitness_2 \leftarrow \textsc{PlayGameOff}(PModel,child_2)$\;
    $pop\leftarrow\textsc{replace}(pop,child_1,child_2)$; \tcp{(popsize + 2) replacement}
    $i \leftarrow i + 1$\;
} \Return{{\rm best solution in }$pop$}\;
\end{algorithm}

\normalsize

Evaluating the fitness of an individual $x$ requires the simulation
(off-line) of the game between the player model and the virtual
player strategy encoded in $x$. The fitness function depends on the
statistical data collected at the end of the simulation. The higher
the number of statistical data to be collected, the higher the
computational cost will be. A priori, a good policy is to consider a
limited number of data. Four data were used in our experiments
during off-line evaluation: A: Number of deaths in the human player
army; B: number of deaths in virtual player army; C: number of
movements; and D: victory degree (i.e., 1 if virtual player wins and
2 otherwise). Fitness function was then defined as $ fitness(x) =
\frac{10000*(A-B)}{C*D}$; higher the fitness value, better the
strategy. This fitness was coded to evolve towards aggressive
solutions.

\section{Experimental analysis}

The experiments were performed using two algorithms: our initial
expert system (RBP), and the algorithm PMEA (i.e., player modeling +
EA) presented in Algorithm \ref{code:improvement}. As to the PMEA,
the EA uses $popsize=50$, $p_X=.7$, $p_M=.01$, and MaxGenerations =
125; mutation is executed as usual at the gene level by changing an
action to any other action randomly chosen. Three different
scenarios where created for experimentation: (1) A map with size $50
\times 50$ grids, 48 agents in VP army, and 32 soldiers in the human
player (HP) team; (2) a map
 $54 \times 46$, with 43 VP soldiers, and 43 HP units; and
(3) a map $50 \times 28$, with 48 VP soldiers, and 53 HP units.
Algorithm \ref{code:improvement} was executed for a value of
$\wp=20$ (i.e., 20 different games were sequentially played), and
the RBP was also executed 20 times; Table \ref{tab:resultsAll1}
shows the results obtained.

\vspace{-3mm}
\begin{table}[!ht]
\small
\caption{Results: $\text{VP}_{win}$ = number of virtual player's
victories, $\text{HP}_{win}$ = number of human player's victories,
$\overline{\text{HP}_{death}}$ = average number of deaths in the HP
army, $\overline{\text{VP}_{death}}$ = average number of deaths in
the VP army, $\overline{mov}$ = average number of movements, and
$\overline{time}$ = average time (minutes) dedicated per
game}
\label{tab:resultsAll1}
\begin{center}
\small
\begin{tabular}{ r  r  r  c  c  c  c  c  r  r}
                     &                     &       & $\text{VP}_{win}$ & $\text{HP}_{win}$ &  $\overline{\text{HP}_{death}}$   &  $\overline{\text{VP}_{death}}$& $\overline{mov}$    &  $\overline{time}$ \\ \hline\hline
$map_{50\times 50}$  & RBP                 &       & 4                 & 16                &   6                               &   7                            & 5345                &   3.56             \\
                     & PMEA                &       & 6                 & 14                &   7                               &   7                            & 4866                &   3.20             \\
\hline
$map_{54\times 46}$  & RBP                 &       & 9                 & 11                &   4                               &   3                            & 7185                &   4.79             \\
                     & PMEA                &       & 7                 & 13                &   6                               &   7                            & 5685                &   3.80             \\
\hline
$map_{50\times 28}$  & RBP                 &       & 3                 & 17                &   3                               &   2                            & 6946                &   4.63             \\
                     & PMEA                &       & 6                 & 14                &   7                               &   6                            & 6056                &   3.78             \\
\hline
\end{tabular}
\end{center}
\end{table}

\vspace{-3mm} Even though in two of the three scenarios PMEA behaves
better than RBP, note that no significant differences are shown;
this is however an expected result as we have considered just one
player what means that the player models obtained in-between two
games are likely similar and thus their corresponding virtual
players also are. In any case, this demonstrates that our approach
is feasible as it produces virtual players comparable - and
sometimes better - to specific and specialized pre-programmed
scripts.

\vspace{-3mm}
\section{Conclusions}

We have described an algorithm to design automatically strategies
exhibiting emergent behaviors that adapt to the user skills in a
one-player war real time strategy game (wRTS); to do so, a model to
mimic how the human player acts in the game is first constructed
during the game, and further a strategy for a virtual player is
evolved (in between games)  via an evolutionary algorithm.

Our proposal was compared with an expert system designed
specifically for the game. Whereas no significance differences have
been highlighted in the experiments, we make note  that our approach
has evident advantages compared to classical manufactured scripts
(i.e., expert systems) used in videogame industry: for instance, it
avoids the predictability of actions to be executed by the virtual
player and thus guarantees to maintain the interest of the player
This is specially interesting when the game involves more
than one player as our approach would allow to construct virtual
players adapted particularly to each of the human players (and this
cannot be obtained with a pre-programmed script).

Further research will cope with multi-player games and thus
multi-objective evolutionary programming techniques should be
considered.

\vspace{-4mm}
\subsubsection*{Acknowledgements}
This work is supported by project NEMESIS (TIN-2008- 05941) of the
Spanish Ministerio de Ciencia e Innovaci\'{o}n, and project TIC-6083
of Junta de Andaluc\'{\i}a.

\vspace{-3mm}
\bibliography{biblio}
\bibliographystyle{splncs}

\end{document}